\documentclass[letterpaper]{article} 
\usepackage{aaai2026}  
\usepackage{times}  
\usepackage{helvet}  
\usepackage{courier}  
\usepackage[hyphens]{url}  
\usepackage{graphicx} 
\def\UrlFont{\rm}  
\usepackage{natbib}  
\usepackage{caption} 
\usepackage{algorithm}
\usepackage{algorithmic}
\nocopyright
\usepackage{newfloat}
\usepackage{listings}
\DeclareCaptionStyle{ruled}{labelfont=normalfont,labelsep=colon,strut=off} 
\floatstyle{ruled}
\newfloat{listing}{tb}{lst}{}
\floatname{listing}{Listing}
\usepackage{comment}
\title{Identifying Evidence-Based Nudges in Biomedical Literature with Large Language Models}
\author{
    Jaydeep Chauhan\textsuperscript{\rm 1},
    Mark Seidman\textsuperscript{\rm 1},
    Pezhman Raeisian Parvari\textsuperscript{\rm 2},
    Zhi Zheng\textsuperscript{\rm 3},
    Zina Ben-Miled\textsuperscript{\rm 4},
    Cristina Barboi\textsuperscript{\rm 1,5},
    Andrew Gonzalez\textsuperscript{\rm 1,5},
    Malaz Boustani\textsuperscript{\rm 1,5,6},
}
\affiliations{
    \textsuperscript{\rm 1} Center for Health Innovation and Implementation Science, Indiana University School of Medicine, Indianapolis, IN, USA \\
    \textsuperscript{\rm 2} Indiana University, Bloomington, IN, USA \\
    \textsuperscript{\rm 3} University of Notre Dame, Notre Dame, IN, USA \\
    \textsuperscript{\rm 4} Phillip M. Drayer Department of Electrical and Computer Engineering, Lamar University, TX, USA \\
    \textsuperscript{\rm 5} Regenstrief Institute, Inc., Indianapolis, IN, USA \\
    \textsuperscript{\rm 6} Sandra Eskanazi Center for Brain Care Innovation, Eskanazi Health, Indianapolis, IN, USA \\
    \{jtchauha, mjseidma, praeisia, mboustan, cbarboi, andrewg\}@iu.edu, \\
    zzheng3@nd.edu, zbenmiled@lamar.edu

}

\usepackage{bibentry}

\begin{document}

\maketitle

\begin{abstract}
We present a scalable, AI-powered system that identifies and extracts evidence-based behavioral nudges from the unstructured biomedical literature to support positive behavioral decision-making in healthcare. Nudges, which are subtle, noncoercive interventions that influence behavior without limiting choice, have demonstrated a strong impact on improving health outcomes (e.g., medication adherence, vaccination uptake). However, identifying these interventions from large scientific corpora remains a bottleneck; PubMed alone contains more than 8 million unstructured articles.

Our system addresses this challenge via a novel multi-stage pipeline. First, we apply hybrid filtering using keyword heuristics, Term Frequency - Inverse Document Frequency (TF-IDF), cosine similarity, and a domain-specific “nudge-term bonus” reducing the corpus to \(\sim\)81,000 candidate articles. In the second stage, we use OpenScholar, a quantized open-source version of LLaMA 3.1 8B fine-tuned on scientific texts, to classify nudge-relevant papers and extract structured fields (e.g., nudge type, cognitive bias, target behavior) in a single forward pass. The outputs are validated against a constrained JSON schema to ensure consistency and interpretability.

We evaluated four inference configurations on a manually labeled test set (N=197). The best-performing set-up (Title/Abstract/Intro) achieved 67.0\% F1 score and 72.0\% recall, ideal for high-recall discovery. A high-precision variant using self-consistency (7 randomized passes with high-temperature sampling) achieved 100\% precision with reduced recall (12\%), demonstrating a tunable trade-off between the exploratory and high-trust use cases. 

This system is currently being validated and integrated into Agile Nudge+, a real-world nudge design platform. Once deployed, it will ground LLM-generated behavioral interventions in peer-reviewed evidence. This work demonstrates a new class of interpretable, domain-specific retrieval pipelines, with practical applications in cross-domain evidence synthesis, meta-analysis acceleration, and personalized healthcare.
\end{abstract}

\section{Introduction}
Behavioral nudges, which are subtle, non-coercive interventions that guide decision making without restricting choice, have demonstrated a measurable impact in healthcare, from improving medication adherence to increasing vaccination uptake \cite{thaler2008nudge, patel2023nudges,boustani2020agile}. During the COVID-19 pandemic, low-cost strategies such as default vaccination appointments and risk-framing messages measurably increased uptake at the population level \cite{banerjee2024default, kasting2019hepb}.

However, the evidence base for effective nudges is buried in the biomedical literature, dispersed across unstructured text, and inconsistently labeled. As of July 2025, PubMed contained approximately 8.1 million full-text articles, which makes manual discovery prohibitively slow for researchers conducting systematic reviews or designing interventions. Keyword search is insufficient, retrieving irrelevant results (for example, 'nudging a neuron') while missing valid nudges described in contextually different terms.

To address this challenge, we present a two-stage AI pipeline that combines classical information retrieval with large-language model (LLM) reasoning to identify and extract evidence-based behavioral nudges from biomedical literature at scale. The system first narrows the search space using a hybrid keyword, TF-IDF, and semantic similarity filtering. It then applies an LLM fine-tuned for scientific text to classify candidate articles using behavioral science inclusion criteria and to extract structured metadata (nudge type, targeted cognitive bias, and intervention goal).

Our key contributions are as follows:

\begin{enumerate}
    \item Scalable nudge discovery — A pipeline capable of processing millions of biomedical articles to surface high-quality behavioral interventions.
    \item LLM-based classification and structured extraction — Integrating precision filtering with interpretable metadata capture in a single pass.
    \item The system is being actively incorporated into Agile Nudge+ \cite{shojaei2024feasibility}, a real-world platform to design personalized behavioral interventions. By grounding its recommendations in a vetted, structured evidence base, the platform can improve reliability, transparency, and trust in AI-generated behavioral recommendations.
\end{enumerate}

To our knowledge, this is the first system to combine large-scale biomedical filtering with behavioral science-aware classification and extraction, creating an evidence base suitable for direct integration into healthcare decision support.

\section{Background and Related Work}

\subsection{Agile Science}

Agile science is a rapidly evolving adaptive process for knowledge discovery and acquisition within the dynamic, constantly changing real-world. The field integrates insights from behavioral economics, complexity science, and 
network science to understand, predict, and steer the 
behaviors of both an individual human and a social 
organization. The practice enables scalable and effective human-centered strategies, processes and tools, including nudges, to be implemented into routine care and subsequently diffused across various social networks \cite{mehta2022conductor}.

\subsection{Behavioral Nudges in Healthcare}

Nudges are rooted in behavioral economics and cognitive psychology, influencing behavioral change by altering the decision environment without limiting freedom of choice \cite{thaler2008nudge}. In healthcare, they have been used to promote preventive screening, improve appointment attendance, and support chronic disease management, often with minimal cost and high scalability. Their success has driven inclusion in public health guidelines and implementation science frameworks.

\subsection{Literature Mining and Information Extraction with LLMs}

Recent advances in LLMs such as GPT \cite{achiam2023gpt4}, BERT \cite{devlin2018bert}, and LLaMA \cite{grattafiori2024llama3} have transformed the literature mining, enabling document classification, summarization, and information extraction at scale. Biomedical NLP has also benefited from domain-specific adaptations such as BioBERT \cite{lee2020biobert} and PubMedBERT \cite{gu2021domain}, which improve named entity recognition and retrieval for scientific texts. Tools such as Elicit and ASReview apply transformer models to accelerate systematic review screening, but these approaches typically focus on general textual relevance rather than on behavioral science constructs central to nudge research, such as choice architecture elements (e.g., defaults, framing, salience, ordering effects) and underlying cognitive biases (e.g., loss aversion, status quo bias, anchoring).

\subsection{Limitations of Existing Tools}

Mining behavioral interventions from the biomedical literature is fundamentally an information retrieval challenge: the goal is to surface a small, high-value subset of documents from a massive corpus, where relevance is defined by domain-specific constructs rather than broad topical similarity. In our case, these constructs include choice architecture (e.g., altering the environment in which decisions are made) and cognitive biases (e.g., loss aversion, present bias) that underlie behavioral nudges. These concepts are often described implicitly, scattered across different sections of a paper, and not captured by standard indexing methods such as keyword matching or MeSH terms.

Traditional tools like PubMed, Cochrane Review Manager, and Publish or Perish excel at retrieving documents based on keyword or citation metrics, but are over- and under-inclusive when the target concepts are abstract, interdisciplinary, or expressed without consistent terminology. Prior biomedical NLP pipelines (e.g., BERT-based screening models) improve general relevance ranking, but rarely incorporate the semantic reasoning and contextual inference needed to detect nudges embedded in complex behavioral or experimental designs.

Our approach addresses this gap by combining large-scale candidate filtering with domain-aware semantic classification and structured metadata extraction, bridging established IR techniques with behavioral science–specific reasoning. This positions the system as both an applied solution for behavioral health and a transferable framework for other specialized domains where fine-grained, concept-level retrieval is essential.

\section{System Overview}
Our system automates the discovery and structuring of evidence-based behavioral nudges from the large-scale biomedical literature, enabling rapid evidence synthesis for healthcare decision making and behavioral change design. It follows a two-stage architecture designed for scalability, precision, and downstream integration.

\subsection{Stage 1 — Hybrid Filtering:}
We first narrowed the search space from millions of PubMed articles to a manageable set of candidates using a hybrid retrieval strategy. This combines:

\begin{itemize}
\item Broad keyword matching to maximize recall.

\item TF-IDF and cosine similarity to capture semantic relevance beyond exact keyword matches.

\item A domain-specific “nudge-term bonus” to upweight documents more likely to describe behavioral interventions.
\end{itemize}
This stage reduces the corpus size by \(\sim\)99\% while maintaining high coverage of potential nudge-relevant studies.

\subsection{Stage 2 — LLM-Based Classification and Structured Extraction:}
The filtered set is passed to an open-source LLaMA 3.1 8B model fine-tuned on the scientific text. In a single forward pass, the model:
\begin{enumerate}

\item Classifies whether the article describes a valid behavioral nudge based on strict inclusion and exclusion criteria rooted in agile science.

\item Extracts structured metadata, including:
\begin{itemize}
\item Nudge type (e.g., default, framing, reminder)

\item Targeted cognitive bias (e.g., loss aversion, social norms)
\end{itemize}
\end{enumerate}

The output is validated against a constrained JSON schema, ensuring interpretable and machine-readable results suitable for direct integration into applications.

From an IR perspective, this stage acts as a semantic reranker, operating on a high-recall candidate set from Stage 1 and applying fine-grained, domain-specific relevance modeling that goes beyond surface lexical similarity. The result is a high-precision, semantically curated subset (\(\sim\)12,000 documents) ready for evidence synthesis, database construction, or retrieval-augmented generation (RAG) applications.

\section{Methods}
\subsection{Hybrid Filtering Stage (TF-IDF + Keyword Bonus)}

Our first stage filter reduces the ~8 million articles in PubMed to a high-recall and manageable subset suitable for computationally expensive downstream classification. This stage was implemented and executed on the Jetstream2 academic cloud platform, which provided the scalability needed to process the entire corpus under resource constraints.
 
\subsection{Data Source and Initial Keyword Expansion}
We started by downloading the PubMed Open Access subset to Jetstream2 storage, containing titles, abstracts, introductions, and metadata in structured JSON format. The initial retrieval step used a curated set of keywords drawn from the behavioral science literature \cite{thaler2008nudge, mehta2023agile}, covering only some common nudge related keywords (e.g. agile science, nudge theory, behavioral interventions). This step retrieved approximately 440,000 articles ($\sim$5.5\% of PubMed), prioritizing the recall over precision.

\subsection{TF-IDF Vectorization and Semantic Matching}
We computed TF-IDF representations using unigrams, bigrams, and trigrams 
(\texttt{ngram\_range}=(1,3)) while excluding extremely rare terms 
(\texttt{min\_df}=2) and overly common terms (\texttt{max\_df}=0.85). 
A domain-specific reference vector was constructed from the expanded set of keywords list covering:

\begin{itemize}
\item Core nudge terms (e.g., nudge, choice architecture, loss aversion),
\item Intervention terms (e.g., randomized, controlled trial, impact),
\item Domain-specific behavioral concepts (e.g., reminders, social proof, present bias).
\end{itemize}
The complete keyword lists are provided in Appendix A to ensure reproducibility. 

Cosine similarity was computed between each article vector and the reference  vector using sparse matrix operations, enabling millions of comparisons to be efficiently executed within the Jetstream2 environment.

\subsection{Keyword Bonus Heuristic}
Although TF-IDF similarity captures broad lexical overlap, it can dilute the impact of rare but highly discriminative domain terms (e.g., 'choice architecture', 'default option', 'loss aversion') that are crucial for identifying behavioral nudges. To address this, we introduce a 'nudge-term bonus' that explicitly increases the retrieval score for documents whose title or abstract contains high-value behavioral science terms.

The bonus is computed as:
\begin{equation}
\mathrm{bonus}(d) = \min\big(|T_d| \times 0.1, \, 0.3\big)
\end{equation}
where $T_d$ is the set of kaywords matching in the document $d$. 

The final \emph{hybrid score} is then:
\begin{equation}
\mathrm{hybrid\_score}(d) = \mathrm{cos\_sim}(d, q) + \mathrm{bonus}(d)
\end{equation}
where $\mathrm{cos\_sim}(d, q)$ is the cosine similarity between the document 
vector $d$ and the reference vector $q$.

Scaling factor 0.1 and cap 0.3 were empirically determined by iterative trial and error on a 100 document development set sampled from PubMed. If the bonus was set too high, the system disproportionately favored documents with explicit nudge-related terminology, leading to precision gains but substantial recall loss (e.g. missing articles that described nudges implicitly). In contrast, setting the bonus too low caused high recall but poor ranking quality, allowing generic biomedical articles with minimal nudge relevance to dominate the top results.

\subsection{Scalability and Filtering Outcome}
This stage was implemented for streaming batch processing with checkpointing, enabling the entire PubMed corpus to be processed in parallelizable chunks 
without exceeding node memory limits. Using a threshold of $0.12$ on the 
hybrid score reduced the candidate pool to approximately 81{,}000 articles 
($\sim$1\% of PubMed), a 98\% corpus size reduction. This ensured that the 
second stage LLM classification could run efficiently while maintaining a broad 
coverage of potential nudge-related literature.

\section{LLM-Based Classification and Extraction}

\subsection{Model Setup and Deployment}

The second stage of our pipeline applies a large language model to the reduced candidate set from Stage~1. 
We use OpenScholar \cite{asai2024openscholar}, an open-source LLaMA 3.1 8B parameter model fine-tuned for scientific literature understanding. 
The model is deployed on a Jetstream2 GPU node (16 vCPUs,
60 GB RAM, 20 GB A100 GPU), using AWQ \cite{lin2024awq} INT4 quantization 
to reduce the footprint of GPU memory  without significant degradation in accuracy. 

\subsubsection{Prompt and Input Structure}
For each retained article, we pass the \emph{title}, \emph{abstract}, and \emph{introduction} into a structured, 
instruction-following prompt directing the model to:
\begin{enumerate}
    \item Classify whether the article describes a \emph{behavioral nudge intervention}.
    \item If relevant, extract the following structured fields:
    \begin{itemize}
        \item \textbf{Nudge type(s)} (e.g., default, framing, reminders)
        \item \textbf{Targeted cognitive bias(es)} (e.g., loss aversion, anchoring)
        \item \textbf{Problem behavior addressed}
        \item \textbf{Target behavior influenced}
        \item \textbf{Reasoning trace} for interpretability
    \end{itemize}
\end{enumerate}
The complete prompt template, including both task instructions and few-shot examples, is provided in Appendix B for transparency and reproducibility. 
The prompt explicitly instructs the model to return its output in a strict 'JSON' format. 
The outputs are validated using Python’s 'json' module, with up to two re-attempts for malformed generations.

Illustrative examples of model outputs for both nudge-related and non-nudge-related articles are included in Appendix C.  

\subsubsection{Inclusion and Exclusion Criteria}
An article is labeled 'positive' if it describes a non-coercive, choice-architectural intervention that alters the digital, social, or physical environment to influence individual behavior without restricting choice and without relying on medical, clinical, or policy-level mechanisms. Qualifying interventions target behaviors such as healthy eating, smoking reduction, physical activity, or mental health improvement, and operate through automatic cognitive processes (e.g., heuristics, biases) rather than only informational or educational means. Appendix B contains the full prompt along with detailed inclusion and exclusion criteria.  
ensure reproducibility.

Few-shot examples embedded in the classification prompt illustrate these boundaries, helping to distinguish valid behavioral nudges from general health education campaigns, medical treatments, or systemic policy changes.

\subsubsection{Self-Consistency Voting}
To improve the reliability of the method, we also evaluated the self-consistency method 
\cite{wang2023selfconsistency, chen2024universal}, in which each document is evaluated 
seven times with high-temperature sampling (T = 0.8). The final decision is obtained by majority vote.
This approach reduces the influence of stochastic model outputs and promotes high-confidence predictions, although it may reduce recall in exchange for greater precision.

\subsubsection{LLM-as-a-Judge Verification}
In selected configurations, we introduce an \emph{LLM-as-a-Judge} step \cite{li2024llmsjudges}, 
where the model re-assesses its own classification in a secondary prompt that explicitly 
asks whether the initial output meets the defined inclusion criteria. This meta-evaluation 
is particularly effective for flagging ambiguous or borderline cases, improving trust and interpretability in the final classification results. The full judge verification prompt is provided in Appendix B.

\section{Experimental Setup}

\subsection{Dataset and Preprocessing}
We used the PubMed database (\(\sim\)8M articles) as our primary corpus.  
The hybrid filtering stage, as described in the previous section, reduced this to \(\sim\)81{,}000 candidate documents (\(\sim\)1\% of PubMed) while preserving broad topical coverage. The title, abstract, and introduction text were extracted for each article; full-text parsing was enabled for select runs.

\subsection{LLM Processing}
The LLaMA 3.1 8B OpenScholar model was applied to all 81{,}000 documents using a continuous batching. A single forward pass jointly:
\begin{enumerate}
    \item Classified articles as nudge-relevant or not.
    \item Extracted structured fields: nudge type, cognitive bias, problem behavior, and target behavior.
\end{enumerate}
This yielded \(\sim\)12{,}000 high-confidence positives for further evaluation.

\subsection{Evaluation Set}
A gold standard set of 197 articles was randomly sampled from the 12{,}000 positives and manually labeled by a domain expert, blinded to model output. Labels followed the same inclusion criteria as the classification prompt:
non-coercive intervention, environmental/choice-architecture design, and exclusion of purely medical or policy mechanisms.
The set contained 86 positives and 111 negatives.

\subsection{Model Configurations}
We tested four configurations with LLM as a judge verification prompt:
\begin{itemize}
    \item \textbf{Title+Abstract+Intro:} LLaMA 3.1 8B, $T=0.1$, single pass.
    \item \textbf{Full Text:} Same model with complete article input.
    \item \textbf{Self-Consistency ($k=7$):} Seven runs at $T=0.8$; label by majority vote:
    \[
    \hat{y} = \arg\max_{c} \sum_{i=1}^k \mathbf{1}[y_i = c]
    \]
    \item \textbf{Gemini 2.5 Pro:} API-based, $T=0.1$, full-text classification.
\end{itemize}

\subsection{Metrics}
We report standard classification metrics—Precision, Recall, F1 score, and Accuracy, to evaluate the performance of our filtering approach. These metrics capture the trade-off between recall and precision, which is particularly important when distinguishing between exploratory analysis and high-trust applications.

\section{Results}
\subsection{Classification Performance}

We tested four model configurations on a manually labeled test set of 197 articles, with the results shown in Table \ref{tab:classification-performance}. The configuration using only the title, abstract, and introduction achieved the highest recall (72\%) and F1 score (67\%), making it the most effective for applications requiring comprehensive coverage, such as literature review phases or exploratory research. Its relatively high accuracy (69\%) and balanced precision (63\%) indicate strong general performance in surfacing a broad range of potentially relevant nudge interventions.

\begin{table}[h]
\centering
\setlength{\tabcolsep}{2pt} 
\begin{tabular}{p{3.0cm}cccc} 
\hline
\textbf{Method} & \textbf{Precision} & \textbf{Recall} & \textbf{F1 Score} & \textbf{Accuracy} \\ \hline
LLaMA 3.1 8B (Title, Abstract, Intro) & 0.63 & 0.72 & 0.67 & 0.69 \\
LLaMA 3.1 8B (Full Document) & 0.72 & 0.51 & 0.60 & 0.70 \\
LLaMA 3.1 8B (Self-Consistency x7) & 1.00 & 0.12 & 0.21 & 0.61 \\
Gemini 2.5 Pro (Full Document) & 0.61 & 0.65 & 0.63 & 0.66 \\ \hline
\end{tabular}
\caption{Performance of classification models on labeled test set ($N = 197$).}
\label{tab:classification-performance}
\end{table}

By contrast, the self-consistency configuration, in which each document was classified seven times and the final label determined by majority vote, delivered perfect precision (100\%) but at the cost of severely reduced recall (12\%). This makes it well-suited for generating gold-standard corpora or other high-trust datasets where false positives must be avoided, but impractical for large-scale discovery. The full-document configuration of LLaMA 3.1 8B improved precision to 72\% compared to the truncated input version, but recall dropped to 51\%, suggesting that additional context sometimes introduces noise or distracts the model from key classification cues. The Gemini 2.5 Pro configuration, which operates on full documents, achieved balanced performance across all metrics, making it a viable alternative in settings where open-source deployment is not feasible.

\subsection{Error Analysis}

Analysis of false positives revealed that models occasionally hallucinated the presence of behavioral interventions based on loosely related descriptions, particularly when interventions were vaguely defined or mentioned primarily in the discussion section. Systematic reviews were another common source of error; despite explicit instructions to include only empirical evaluations of nudge interventions, the structured and concept-rich format of reviews often led to misclassification.

False negatives most often resulted from non-standard terminology or unconventional framing of nudges, which prevented the model from recognizing legitimate interventions. Some cases involved interventions embedded within broader clinical or population-level programs, making it difficult for the model to isolate and classify the behavioral component. Structured information extraction generally succeeded for primary fields such as nudge type and cognitive bias, but consistency decreased for complex studies involving multiple nudges or multiple experimental arms.

We also observed that the LLM-as-a-judge mechanism, used to verify classifications, was highly sensitive to prompt wording and occasionally biased toward over-conservative decisions. The self-consistency approach, while effective in eliminating false positives, came with high computational cost and low throughput, making it unsuitable for large-scale application. In some cases, both verification strategies appeared to produce post hoc justifications for incorrect predictions rather than genuine reasoning traces, a phenomenon of unfaithfulness observed in prior work \cite{turpin2024unfaithful}. Finally, as noted in prior studies \cite{liu2023lost, kandpal2022longtail}, providing full-text input can sometimes reduce model performance compared to truncated input, likely because longer documents contain distracting or redundant content that diminishes the model’s focus on relevant evidence.
 
Detailed examples of both false positive and false negative cases, illustrating the types of errors described above, are provided in Appendix D.

\section{Discussion}

\subsection{Applied Impact for Evidence-Based Healthcare}

The results demonstrate that large language models, when combined with a scalable two-stage filtering pipeline, can efficiently identify behavioral nudge interventions within massive biomedical corpora. Our system reduced the PubMed database of more than eight million articles to approximately 12,000 high-relevance studies, offering a valuable resource for healthcare strategy design, rapid evidence synthesis, and integration into decision-support tools. The curated corpus is being incorporated into Agile Nudge+. This integration will enable retrieval-augmented generation, grounding AI-generated recommendations in verified literature to enhance both transparency and trustworthiness.

\subsection{Limitations and Risks}

Despite these benefits, there are limitations. Large language models are prone to both false positives and false negatives, especially when confronted with atypical terminology or ambiguous descriptions of interventions. Overly strict inclusion criteria, as seen in the self-consistency and judge-mode configurations, can limit recall and exclude borderline but relevant studies. Furthermore, full-text processing can dilute model performance, suggesting that targeted context windows may be more effective for classification. Computational costs, particularly for high-precision modes, also limit scalability in resource-constrained environments.

Ethical risks must be considered when automating the discovery of behavioral interventions. Misclassifications could propagate misleading evidence into downstream systems, and reliance on automated classification without expert oversight risks bias amplification. 

\subsection{Ethical Considerations}

Automating the identification and recommendation of behavioral interventions introduces ethical concerns. There is a risk of amplifying the biases present in the training data, misrepresenting interventions, or overrelying on model outputs without sufficient expert oversight. To mitigate these concerns, LLM-generated output will be subjected to downstream validation layers, including human-in-the-loop review, transparency in traceable reasoning, and strict adherence to inclusion criteria rooted in behavioral science.

Ultimately, this system represents a step towards responsible and transparent AI-assisted behavioral intervention design, offering an important tool for evidence-based healthcare and digital policy development.

\section{Conclusion}

We presented a scalable, fully automated pipeline for identifying and classifying evidence-based behavioral nudges from the biomedical literature. The system integrates a two-phase filtering approach -- combining TF-IDF and cosine similarity with LLM-based classification — to reduce the PubMed corpus of more than eight million articles to approximately 12,000 highly relevant studies. Evaluation across multiple model configurations revealed clear precision–recall trade-offs, offering flexible modes for different use cases. The LLaMA 3.1 8B model using only titles, abstracts, and introductions achieved the highest recall and F1 score, making it optimal for broad discovery, while the self-consistency mode achieved perfect precision for high-trust applications. This corpus will be integrated into Agile Nudge+ to enhance the reliability and transparency of AI-generated nudge recommendations, forming the basis for a Retrieval-Augmented Generation (RAG) framework that links user-specified behaviors with verified, structured evidence.

\bibliography{aaai2026}

\begin{thebibliography}{21}
\providecommand{\natexlab}[1]{#1}

\bibitem[{Achiam et~al.(2023)Achiam, Adler, Agarwal, Ahmad, Akkaya, Aleman, Almeida, Altenschmidt, Altman, Anadkat, and et~al.}]{achiam2023gpt4}
Achiam, J.; Adler, S.; Agarwal, S.; Ahmad, L.; Akkaya, I.; Aleman, F.~L.; Almeida, D.; Altenschmidt, J.; Altman, S.; Anadkat, S.; and et~al. 2023.
\newblock GPT-4 Technical Report.
\newblock \emph{arXiv preprint arXiv:2303.08774}.

\bibitem[{Asai et~al.(2024)Asai, He, Shao, Shi, Singh, Chang, Lo, Soldaini, Feldman, D’arcy et~al.}]{asai2024openscholar}
Asai, A.; He, J.; Shao, R.; Shi, W.; Singh, A.; Chang, J.~C.; Lo, K.; Soldaini, L.; Feldman, S.; D’arcy, M.; et~al. 2024.
\newblock OpenScholar: Synthesizing Scientific Literature with Retrieval-Augmented LMs.
\newblock \emph{arXiv preprint arXiv:2411.14199}.

\bibitem[{Banerjee et~al.(2024)Banerjee, John, Nyhan, Hunter, Koenig, Lee-Whiting, Loewen, McAndrews, and Savani}]{banerjee2024default}
Banerjee, S.; John, P.; Nyhan, B.; Hunter, A.; Koenig, R.; Lee-Whiting, B.; Loewen, P.~J.; McAndrews, J.; and Savani, M. 2024.
\newblock Thinking about Default Enrollment Lowers Vaccination Intentions and Public Support in G7 Countries.
\newblock \emph{PNAS Nexus}, page093.

\bibitem[{Boustani, Azar, and Solid(2020)}]{boustani2020agile}
Boustani, M.; Azar, J.; and Solid, C. 2020.
\newblock \emph{Agile Implementation: A Model for Implementing Evidence-Based Healthcare Solutions into Real-world Practice to Achieve Sustainable Change}.
\newblock Morgan James.

\bibitem[{Chen et~al.(2024)Chen, Aksitov, Alon, Ren, Xiao, Yin, Prakash, Sutton, Wang, and Zhou}]{chen2024universal}
Chen, X.; Aksitov, R.; Alon, U.; Ren, J.; Xiao, K.; Yin, P.; Prakash, S.; Sutton, C.; Wang, X.; and Zhou, D. 2024.
\newblock Universal self-consistency for large language models.
\newblock In \emph{ICML 2024 Workshop on In-Context Learning}.

\bibitem[{Devlin et~al.(2018)Devlin, Chang, Lee, and Toutanova}]{devlin2018bert}
Devlin, J.; Chang, M.-W.; Lee, K.; and Toutanova, K. 2018.
\newblock {BERT}: Pre-training of Deep Bidirectional Transformers for Language Understanding.
\newblock \emph{arXiv preprint arXiv:1810.04805}.

\bibitem[{Grattafiori et~al.(2024)Grattafiori, Dubey, Jauhri, Pandey, Kadian, Al-Dahle, Letman, Mathur, Schelten, Vaughan et~al.}]{grattafiori2024llama3}
Grattafiori, A.; Dubey, A.; Jauhri, A.; Pandey, A.; Kadian, A.; Al-Dahle, A.; Letman, A.; Mathur, A.; Schelten, A.; Vaughan, A.; et~al. 2024.
\newblock The LLaMA 3 Herd of Models.
\newblock \emph{arXiv preprint arXiv:2407.21783}.

\bibitem[{Gu et~al.(2021)Gu, Tinn, Cheng, Lucas, Usuyama, Liu, Naumann, Gao, and Poon}]{gu2021domain}
Gu, Y.; Tinn, R.; Cheng, H.; Lucas, M.; Usuyama, N.; Liu, X.; Naumann, T.; Gao, J.; and Poon, H. 2021.
\newblock Domain-specific language model pretraining for biomedical natural language processing.
\newblock \emph{ACM Transactions on Computing for Healthcare (HEALTH)}, 3(1): 1--23.

\bibitem[{Kandpal et~al.(2022)Kandpal, Deng, Roberts, Wallace, and Raffel}]{kandpal2022longtail}
Kandpal, N.; Deng, H.; Roberts, A.; Wallace, E.; and Raffel, C. 2022.
\newblock Large language models struggle to learn long-tail knowledge.
\newblock \emph{arXiv preprint arXiv:2211.08411}.

\bibitem[{Kasting et~al.(2019)Kasting, Head, Cox, Cox, and Zimet}]{kasting2019hepb}
Kasting, M.~L.; Head, K.~J.; Cox, D.; Cox, A.~D.; and Zimet, G.~D. 2019.
\newblock The effects of message framing and healthcare provider recommendation on adult hepatitis B vaccination: a randomized controlled trial.
\newblock \emph{Preventive Medicine}, 127: 105798.

\bibitem[{Lee et~al.(2020)Lee, Yoon, Kim, Kim, So, and Kang}]{lee2020biobert}
Lee, J.; Yoon, W.; Kim, S.; Kim, D.; So, C.~H.; and Kang, J. 2020.
\newblock BioBERT: a pre-trained biomedical language representation model for biomedical text mining.
\newblock \emph{Bioinformatics}, 36(4): 1234--1240.

\bibitem[{Li et~al.(2024)}]{li2024llmsjudges}
Li, H.; et~al. 2024.
\newblock LLMs-as-Judges: A Comprehensive Survey on LLM-based Evaluation Methods.
\newblock \emph{arXiv preprint arXiv:2412.05579}.

\bibitem[{Lin et~al.(2024)Lin, Tang, Tang, Yang, Chen, Wang, Xiao, Dang, Gan, and Han}]{lin2024awq}
Lin, J.; Tang, J.; Tang, H.; Yang, S.; Chen, W.-M.; Wang, W.-C.; Xiao, G.; Dang, X.; Gan, C.; and Han, S. 2024.
\newblock {AWQ}: Activation-aware weight quantization for on-device {LLM} compression and acceleration.
\newblock In \emph{Proceedings of Machine Learning and Systems}, volume~6, 87--100.

\bibitem[{Liu et~al.(2023)Liu, Lin, Hewitt, Paranjape, Bevilacqua, Petroni, and Liang}]{liu2023lost}
Liu, N.~F.; Lin, K.; Hewitt, J.; Paranjape, A.; Bevilacqua, M.; Petroni, F.; and Liang, P. 2023.
\newblock Lost in the middle: How language models use long contexts.
\newblock \emph{arXiv preprint arXiv:2307.03172}.

\bibitem[{Mehta et~al.(2022)Mehta, Aalsma, O'brien, Boyer, Ahmed, Summanwar, and Boustani}]{mehta2022conductor}
Mehta, J.; Aalsma, M.~C.; O'brien, A.; Boyer, T.~J.; Ahmed, R.~A.; Summanwar, D.; and Boustani, M. 2022.
\newblock Becoming an Agile Change Conductor.
\newblock \emph{Front Publick Health}.

\bibitem[{Mehta et~al.(2023)Mehta, Williams, Holden, Taylor, Fowler, and Boustani}]{mehta2023agile}
Mehta, J.; Williams, C.; Holden, R.~J.; Taylor, B.; Fowler, N.~R.; and Boustani, M. 2023.
\newblock The methodology of the Agile nudge university.
\newblock \emph{Frontiers in Health Services}, 3: 1212787.

\bibitem[{Patel et~al.(2023)Patel, Milkman, Gandhi, Graci, Gromet, Ho, Kay, Lee, Rothschild, Akinola et~al.}]{patel2023nudges}
Patel, M.~S.; Milkman, K.~L.; Gandhi, L.; Graci, H.~N.; Gromet, D.; Ho, H.; Kay, S.~J.; Lee, T.~W.; Rothschild, J.; Akinola, M.; et~al. 2023.
\newblock A randomized trial of behavioral nudges delivered through text messages to increase influenza vaccination among patients with an upcoming primary care visit.
\newblock \emph{American Journal of Health Promotion}, 37: 324--332.

\bibitem[{Shojaei et~al.(2024)Shojaei, Shojaei, Desai, Long, Mehta, Fowler et~al.}]{shojaei2024feasibility}
Shojaei, F.; Shojaei, F.; Desai, A.~P.; Long, E.; Mehta, J.; Fowler, N.~R.; et~al. 2024.
\newblock The feasibility of AgileNudge+ software to facilitate positive behavioral change: a mixed methods design.
\newblock \emph{JMIR Formative Research}.

\bibitem[{Thaler and Sunstein(2008)}]{thaler2008nudge}
Thaler, R.~H.; and Sunstein, C.~R. 2008.
\newblock \emph{Nudge: Improving Decisions About Health, Wealth, and Happiness}.
\newblock Penguin Books.

\bibitem[{Turpin et~al.(2024)Turpin, Michael, Perez, and Bowman}]{turpin2024unfaithful}
Turpin, M.; Michael, J.; Perez, E.; and Bowman, S. 2024.
\newblock Language models don’t always say what they think: unfaithful explanations in chain-of-thought prompting.
\newblock In \emph{Advances in Neural Information Processing Systems}, volume~36.

\bibitem[{Wang et~al.(2023)Wang, Wei, Schuurmans, Le, Chi, Narang, Chowdhery, and Zhou}]{wang2023selfconsistency}
Wang, X.; Wei, J.; Schuurmans, D.; Le, Q.~V.; Chi, E.~H.; Narang, S.; Chowdhery, A.; and Zhou, D. 2023.
\newblock Self-consistency improves chain of thought reasoning in language models.
\newblock In \emph{Proceedings of the 11th International Conference on Learning Representations (ICLR)}.

\end{thebibliography}
\appendix

\section{Appendix A: TF-IDF Keyword Lists}

Below are the complete keyword sets used for the TF-IDF filtering stage.

\subsection{Nudge Keywords}
\begin{lstlisting}
nudge, nudge theory, behavioral intervention, choice architecture,
decision making, behavioral economics, behavior change,
behavioral insights, default option, opt out, opt in,
social norm, anchor, anchoring, framing effect,
choice design, behavioral science, decision architecture,
cognitive bias, heuristic, behavioral design, loss aversion,
incentive, priming, salience, decision fatigue, status quo bias,
behavioral insight, behavioral science, choice design, choice framing
\end{lstlisting}

\subsection{Intervention-Related Keywords}
\begin{lstlisting}
randomized, controlled trial, intervention, experiment,
study design, implementation, evaluation, outcome,
effectiveness, efficacy, impact, effect
\end{lstlisting}

\subsection{Domain-Specific Keywords}
\begin{lstlisting}
incentives, message framing, reminders, salience,
simplification, defaults, commitment devices, priming,
ego, messenger effect, social proof, loss aversion,
public policy, health behavior, financial decision,
environmental choice, pro-social behavior, persuasive design,
choice overload, scarcity, present bias, feedback,
transparency, disclosure, evidence-based policy
\end{lstlisting}

\section{Appendix B: LLM Prompts}

\subsection{Main Prompt}
\begin{lstlisting}[language=Python]
main_prompt: str = (
    """
You are an expert in behavioral science and AI-driven text analysis. Your task is to analyze research papers and extract structured information about **evidenced-based nudge interventions**.

### **Step 1: Classify the Paper**
Determine if the paper describes an **evidence-based nudge intervention** based on the following criteria:

**Inclusion Criteria (ALL must be met for a paper to be classified as "Nudge-Related"):\n**:
1. The study describes an **intervention** that modifies **choice architecture** rather than simply informing or educating.  
2. The intervention **targets individual behavior change** in areas such as:  
   - Improving sleep schedules  
   - Reducing smoking, alcohol consumption  
   - Combating obesity, unhealthy eating  
   - Addressing mental health challenges  
   - Encouraging healthy lifestyle changes  
3. The intervention is **non-coercive, voluntary, and does not involve financial rewards or punishments**.  
4. The nudge mechanism aligns with at least one recognized nudge types (see Step 2).  

**Exclusion Criteria (If ANY are met, classify as "Not Nudge-Related")**:
1. The study describes **a medical treatment, drug, or clinical procedure**.  
2. The intervention relies on **financial incentives, taxes, penalties, or subsidies**.  
3. The study focuses on **policy, law, or government-level interventions** rather than individual choice modification.  
4. The study is a **review article or theoretical discussion** without an actual intervention.  
5. The study does **not describe a specific intervention applied to participants**.  

---

### **Step 2: Extract Information**
If the paper is **nudge-related**, extract the following structured details:

1. **Nudge Types** -> Select one or more relevant nudge mechanisms from this list:  
   - **Messenger** -> Influences based on who delivers the information.  
   - **Incentives** -> Leverages loss aversion but without monetary rewards/punishments.  
   - **Norms** -> Uses social proof (e.g., "90% of people do X").  
   - **Default** -> Pre-selects an option to encourage a behavior.  
   - **Salience** -> Highlights information to make it more noticeable.  
   - **Priming** -> Uses subconscious cues (e.g., healthy food placement).  
   - **Affect** -> Evokes emotional responses to drive action.  
   - **Commitment** -> Encourages self-imposed goals.  
   - **Ego** -> Aligns behavior with self-identity.  

2. **Associated Cognitive Biases** -> Identify biases influencing the intervention (e.g., loss aversion, status quo bias, anchoring).  

3. **Problem Behavior** -> The undesired behavior targeted by the intervention.  

4. **Target Behavior** -> The intended behavioral outcome.  

---

### **Step 3: Final Decision:\n**
- If Step 2 fails to extract relevant details, the paper should be classified as not-nudge related and excluded
- Otherwise, confirm it as **Nudge-Related**

### **JSON Output Format**
Your response must be in **valid JSON format** with the following structure:

```json
{{
"is_nudge_related": true/false,  
"nudge_types": [],
"cognitive_biases": [],
"problem_behavior": "",
"target_behavior": "",
"Reason": ""
}}
```
Do not generate unnecessary explanations beyond this structured output.

### **Input:**
**Title**: {title}  
**Abstract**: {abstract}  
**Introduction**: {intro} 
""".strip()
)
\end{lstlisting}

\subsection{Verification Prompt}
\begin{lstlisting}[language=Python]
verify_nudge_cls_prompt: str = (
"""
You are an expert in behavioral science and AI-assisted systematic reviews. Your task is to verify the following claim about a research paper.

Carefully analyze the paper's **title**, **abstract**, and **full text**, and assess whether the claim is **supported by the content**. If the claim is accurate, respond with **"Yes"**. If the claim is not supported or contradicted by the text, respond with **"No"**.

### The claim:
This paper **{claim}**.

### Use the following criteria to evaluate the claim:

#### Inclusion Criteria (ALL must be met to classify as "Nudge-Related"):
1. The study describes an **intervention** that modifies **choice architecture** altering the digital, social, or physical environment to facilitate behavior change **without restricting choice**.
2. The intervention targets **individual behavior change**, such as:
   - Improving sleep  
   - Reducing smoking or alcohol use  
   - Encouraging healthier eating or weight loss  
   - Addressing mental health  
   - Promoting a healthier lifestyle
3. The intervention is **non-coercive** (no force, mandates, or excessive financial rewards or penalties).
4. The behavior change mechanism relies on **automatic, subconscious cognitive processes**, such as heuristics or biases.
5. The intervention aligns with at least one known **nudge type** (e.g., defaults, salience, reminders, social norms, framing).
6. Has expirical evidence from actual implementation or experiment

#### Exclusion Criteria (ANY of these makes the claim false):
1. The study focuses on **medical treatments**, **drugs**, or **clinical procedures**.
2. The intervention is based on **financial incentives**, **penalties**, **subsidies**, or **taxes**.
3. The focus is on **policy or systemic interventions**, not individual-level behavior.
4. The paper is a **review**, **meta-analysis**, or **theoretical piece** with no implemented intervention.
5. There is **no specific intervention** described or applied to participants.

### Paper:
**Title**: {title}  
**Abstract**: {abstract}  
**Full Text**: {full_text}

### Give your answer in the following format:
{{
"Reasoning": "give your reasoning here explaining your decision"
"Supported": [Yes/No]
}}
""".strip()
)
\end{lstlisting}

\section*{Appendix C: Sample Model Outputs}

To illustrate how the classification and extraction pipeline operates in practice, we provide sample outputs generated by the LLaMA 3.1 8B model. These examples demonstrate both positive and negative classifications, highlighting the structured JSON schema returned by the system.

\subsection*{Identified Articles with Nudge Interventions}
\subsection{Example 1}
\textbf{Paper title:} \emph{A Multisite Randomized Trial of a Motivational Intervention Targeting Multiple Risks in Youth Living with HIV: Initial effects on Motivation, Self-Efficacy and Depression}
\begin{lstlisting}[language=]
{
  "is_nudge_related": true,
  "nudge_types": ["Messenger", "Commitment"],
  "cognitive_biases": ["Loss aversion"],
  "problem_behavior": "Multiple risk behaviors (HIV medication adherence, sexual risk behavior, and substance use)",
  "target_behavior": "Improved motivation, self-efficacy, and reduced depression",
  "Reason": "The study describes an intervention that modifies choice architecture by using motivational interviewing (MI) to increase motivational readiness to change, build self-efficacy, and reduce psychological distress."
}
\end{lstlisting}

\subsection*{Example 2}
\textbf{Paper title:} \emph{ProsCan for Men: Randomised controlled trial of a decision support intervention for men with localised prostate cancer}

\begin{lstlisting}[language=]
{
  "is_nudge_related": true,
  "nudge_types": ["Messenger", "Affect"],
  "cognitive_biases": ["Loss aversion", "Status quo bias"],
  "problem_behavior": "Decision-related distress",
  "target_behavior": "Improved decision-making and adjustment to cancer",
  "Reason": "The study describes a decision support intervention that targets individual behavior change, uses messenger and affect nudge types, and addresses cognitive biases such as loss aversion and status quo bias."
}
\end{lstlisting}
\subsection{Articles Identified as Not Meeting Nudge Criteria}
\subsection{Example 3}
\textbf{Paper title:} \emph{Clinician and cancer patient views on patient participation in treatment decision-making: a quantitative and qualitative exploration}
\begin{lstlisting}[language=]
{
  "is_nudge_related": false,
  "Reason": "The study does not describe an intervention that modifies choice architecture, nor does it target individual behavior change. The study is a qualitative and quantitative exploration of clinician and cancer patient views on treatment decision-making, without an actual intervention."
}
\end{lstlisting}

\subsection*{Example 4}
\textbf{Paper title:} \emph{Olfactory mechanisms of stereotyped behavior: on the scent of specialized circuits}

\begin{lstlisting}[language=]
{
  "is_nudge_related": false,
  "nudge_types": [],
  "cognitive_biases": [],
  "problem_behavior": "",
  "target_behavior": "",
  "Reason": "The paper does not describe an evidence-based nudge intervention. It is a review article discussing mechanisms of olfactory behavior and identification of specialized olfactory circuits."
}
\end{lstlisting}

\section*{Appendix D: Illustrative Error Cases}

\subsection*{False Positives}

\subsection*{Example 1 -- Misclassified Systematic Review}
\textbf{Paper Title}: \textit{Can nudge-interventions address health service overuse and underuse? Protocol for a systematic review} \\
\textbf{Ground Truth}: Not a nudge intervention study (systematic review protocol). \\
\textbf{Model Configuration}: Title + Abstract + Introduction. \\
\textbf{Model Prediction}: Classified as nudge study. \\
\textbf{Model Reasoning (excerpt)}: \\
\begin{quote}
The paper describes nudge-interventions aimed at health professionals to reduce overuse and underuse of health services. The interventions modify choice architecture, target individual behavior change, are non-coercive, voluntary, and do not involve excessive financial rewards or punishments. They attempt to predictably influence health professionals' behavior by targeting subconscious routines and biases present in decision-making. The paper mentions various nudge types, including default, framing, salience, norms, and commitment, and cognitive biases such as default bias, framing effects, loss aversion, order effects, salience effect, and norms. The target behaviors include reducing inappropriate prescribing of antibiotics, reducing inappropriate total knee replacement, reducing inappropriate surgery for low back pain, reducing unnecessary screening tests for cancer, improving prescribing of generic medications instead of brand name medications, improving anticoagulation in patients with atrial fibrillation, improving beta blockers for patients who have had a myocardial infarction, improving exercise prescription for chronic conditions, and reducing wasteful misallocation of resources.
\end{quote}
\textbf{Error Type}: Misclassification of a systematic review. \\
\textbf{Notes}: Also illustrates \textit{post hoc justification} --- the model fabricated detailed reasoning to match its incorrect label.

\subsection*{Example 2 -- Misclassified Theoretical Study}
\textbf{Paper Title}: \textit{Designing an Internet Intervention for Emerging Adults Who Experience Troubled Relationships} \\
\textbf{Ground Truth}: Not a nudge intervention (conceptual/theoretical design). \\
\textbf{Model Configuration}: Title + Abstract + Introduction. \\
\textbf{Model Prediction}: Classified as nudge study. \\
\textbf{Model Reasoning (excerpt)}: \\
\begin{quote}
The paper describes an intervention that modifies choice architecture by targeting individual behavior change in emerging adults experiencing troubled intimate partner relationships. The intervention uses narrative therapy practices as mechanisms of change and employs a human-supported approach with peer counselors to provide personalized responses to users' stories. The target behaviors are specific and actionable, such as writing stories that explore and challenge problematic relational schemas, engaging with the WISER program, and participating in a human-supported approach. The intervention aligns with recognized nudge types, including messenger and default, and leverages cognitive biases such as loss aversion and framing.
\end{quote}
\textbf{Error Type}: False positive due to hallucinated nudge framing. \\
\textbf{Notes}: Shows \textit{unfaithful reasoning traces} --- model generated a nudge framing where none existed.

\subsection*{Example 3 -- System-Level Initiative Misclassified as Nudge}

\textbf{Paper Title}: \textit{Creating the Exceptional Patient Experience in One Academic Health System} \\
\textbf{Ground Truth}: Not a nudge intervention at the individual level (organizational/system-level initiative). \\
\textbf{Model Configuration}: Title + Abstract + Introduction. \\
\textbf{Model Prediction}: Classified as nudge study. \\
\textbf{Model Reasoning (excerpt)}: 
\begin{quote}
The University of Utah Health Care system's Exceptional Patient Experience (EPE) initiative uses various nudge types to improve patient satisfaction and experience. The initiative includes posting patient satisfaction scores and comments on the university's public Web site, which serves as a messenger nudge. The system also uses default nudge by pre-selecting an option to improve patient satisfaction. The initiative also uses loss aversion and framing cognitive biases to influence behavior. The target behaviors include posting patient satisfaction scores and comments, sharing patient feedback and best practices, providing values-based employment and retention, reward, and recognition, improving communication and empathy between patients and providers, and enhancing professionalism and reducing variability in performance.
\end{quote}
\textbf{Error Type}: False positive due to misclassification of a system-level initiative as an individual-level nudge. \\
\textbf{Notes}: The nudge described is not applicable at the individual level, but the model inferred detailed individual-level effects.

\subsection*{Example 4 -- Population-Level Policy Misclassified as Nudge}

\textbf{Paper Title}: \textit{Dynamics of Mask Use as a Prevention Strategy against SARS-CoV-2 in Panama}

\textbf{Ground Truth}: Not a nudge intervention at the individual level (population-level policy).  

\textbf{Model Configuration}: Title + Abstract + Introduction.  

\textbf{Model Prediction}: Classified as nudge study.  

\textbf{Model Reasoning (excerpt)}:
\begin{quote}
The study describes an intervention that modifies choice architecture by promoting the use of surgical masks over cloth masks and encouraging mask usage in suburban neighborhoods through targeted messaging. The intervention targets individual behavior change by reducing the transmission of SARS-CoV-2. The study does not rely on direct financial incentives, taxes, penalties, or subsidies. The intervention aligns with the nudge type 'Messenger' and 'Salience' and attempts to predictably influence behavior by targeting subconscious routines and biases present in decision-making. The study focuses on individual behavior modification rather than policy or government-level interventions.
\end{quote}

\textbf{Error Type}: False positive due to misclassification of a population-level policy as an individual-level nudge.  

\textbf{Notes}: The intervention is primarily a public health policy; the model incorrectly framed it as an individual nudge.

\subsection{False Negatives}
\subsection*{Example 5 – Digital Environment Gamification Misclassified as Nudge}
\textbf{Paper Title:} Competitive gamification in crowdsourcing-based contextual-aware recommender systems

\textbf{Ground Truth:} Nudge intervention (targets engagement behavior via gamification).

\textbf{Model Configuration:} Gemini 2.5 pro with full document

\textbf{Model Prediction:} Classified as non nudge study.

\textbf{Model Reasoning (excerpt):} 
\begin{quote}
The paper describes an intervention using competitive gamification to encourage users to perform crowdsourcing tasks for a restaurant recommender system during COVID-19. While the system's output (restaurant recommendations) could indirectly relate to health by helping users avoid risky places, the intervention itself (gamification) is applied to the crowdsourcers to influence their task engagement behavior, not to the end-users to influence their health-related choices. Specifically, Inclusion Criterion 2 states: The intervention targets individual behavior change, such as: Improving sleep, Reducing smoking or alcohol use, Encouraging healthier eating or weight loss, Addressing mental health, Promoting a healthier lifestyle. The behavior targeted by the gamification intervention in this paper is users performing crowdsourcing tasks and engage more in contributing data. This behavior, while valuable for the system, does not directly align with the examples of individual health or lifestyle behaviors listed in Criterion 2. Therefore, Inclusion Criterion 2 is not met, making the claim unsupported.
\end{quote}
\textbf{Error Type:} False negative due to conservative interpretation of digital gamification as a non-health-related nudge.

\textbf{Notes:} Model was over-conservative; sensitive to prompt wording and classified the digital gamification intervention as non-nudge despite it influencing individual engagement behavior. 

\subsection*{Example 6 – Digital Environment Salience and Priming Misclassified as Non-Nudge}
\textbf{Paper Title:} [Efficacy and Moderation of Mobile App-Based Programs for Mindfulness-Based Training, Self-Compassion Training, and Cognitive Behavioral Psychoeducation on Mental Health: Randomized Controlled Noninferiority Trial]

\textbf{Ground Truth:} Nudge intervention (digital environment salience and priming).

\textbf{Model Configuration:} Gemini 2.5 Pro with full document

\textbf{Model Prediction:} Classified as non-nudge study.

\textbf{Model Reasoning (excerpt):} 
The paper describes mobile app-based programs involving mindfulness-based training, self-compassion training, and cognitive behavioral psychoeducation. While the intervention targets individual mental health and uses digital prompts, the core content involves conscious learning, reflection, and therapeutic exercises rather than automatic, subconscious processes. The intervention does not align with recognized nudge types such as defaults, salience, reminders, social norms, or framing. Although the study is an RCT with empirical evidence, it also meets an exclusion criterion as the interventions are established psychological treatments in a clinical/therapeutic context.

\textbf{Error Type:} False negative due to conservative interpretation of therapeutic app-based interventions as non-nudge, despite features of digital salience and priming.

\textbf{Notes:} The model is highly sensitive to prompt wording and showing over-conservative behavior in identifying digital environment salience and priming interventions.

\subsection*{Example 7 – Digital Engagement Misclassified as Non-Nudge}
\textbf{Paper Title:} Understanding rural women's preferences for telephone call engagement with primary health care providers in Nigeria: a discrete choice experiment

\textbf{Ground Truth:} Nudge intervention (behavioral intervention modifying digital engagement environment).

\textbf{Model Configuration:} Llama 3.1 8B with full document

\textbf{Model Prediction:} Classified as non-nudge study.

\textbf{Model Reasoning (excerpt):} 
The paper describes an intervention that modifies the digital environment to facilitate behavior change by providing telephone call engagement with primary health care providers. While the intervention facilitates behavior change, it does not alter choice architecture in a subtle way without restricting choice, which is a key characteristic of classical nudges. Instead, it provides a service based on user preferences. The study focuses on understanding these preferences rather than shaping behavior via environmental modifications. Therefore, the model judged this as not an evidence-based nudge intervention.

\textbf{Error Type:} False negative due to non-standard terminology and unconventional framing of the intervention.

\textbf{Notes:} The model failed to identify the intervention as a nudge due to unconventional terminology

\subsubsection*{Example 8 – Decision Board Intervention Misclassified as Non-Nudge}
\textbf{Paper Title:} Developing and pre-testing a decision board to facilitate informed choice about delivery approach in uncomplicated pregnancy

\textbf{Ground Truth:} Nudge intervention (physical environment priming).

\textbf{Model Configuration:} Llama 3.1 8B with full text

\textbf{Model Prediction:} Classified as non-nudge study.

\textbf{Model Reasoning (excerpt):} 
The paper describes an intervention that modifies choice architecture by providing systematic evidence-based information to facilitate informed decision-making about delivery approach in uncomplicated pregnancy. The intervention targets individual behavior change, specifically enhancing informed decision-making related to delivery approach. The intervention is non-coercive, as it does not restrict choice or impose penalties. The behavior change mechanism relies on automatic, subconscious cognitive processes, such as heuristics or biases, as the decision board aims to present medical risks and benefits in a clear and systematic way. The intervention aligns with the nudge type of default or framing, as it presents information in a way that influences the decision-making process. The study provides empirical evidence from actual implementation or experiment, as it reports the results of a pre-test of the decision board with 40 women. However, the focus is on policy, not individual-level behavior, as the goal is to inform clinical discussions and enhance informed decision-making related to delivery approach in uncomplicated pregnancy. Therefore, the claim that the paper is an evidence-based nudge intervention is not fully supported.

\textbf{Error Type:} False negative due to the intervention being part of a broader clinical or population-level program, which made it difficult for the model to isolate and classify the behavioral component.

\textbf{Notes:} The decision board uses physical environment priming to guide decision-making, but the model failed to recognize it as a nudge due to its embedding within a clinical setting.

\end{document}